\documentclass{article}

\PassOptionsToPackage{numbers,compress}{natbib}
\usepackage[preprint]{neurips_2026}

\usepackage[utf8]{inputenc} % allow utf-8 input
\usepackage[T1]{fontenc}    % use 8-bit T1 fonts
\usepackage{hyperref}       % hyperlinks
\hypersetup{
    colorlinks=true,      % 开启文字颜色，关闭方框
    linkcolor=blue,       % 内部链接颜色（如图表引用）
    citecolor=blue,       % 文献引用颜色
    urlcolor=blue         % URL 链接颜色
}
\usepackage{url}            % simple URL typesetting
\usepackage{booktabs}       % professional-quality tables
\usepackage{amsfonts}       % blackboard math symbols
\usepackage{nicefrac}       % compact symbols for 1/2, etc.
\usepackage{microtype}      % microtypography
\usepackage{xcolor}         % colors
\usepackage{amsmath,amssymb}
\usepackage{graphicx}
\usepackage{multirow}
\usepackage{algpseudocode}

\title{RISE: Reliable Improvement in Self-Evolving Vision-Language Models}

\author{%
  Chaoran Xu \quad Yingmao Miao \quad Pengfei Zhang\thanks{Project leader.} \quad Hao Dou \quad Lei Sun \quad Xiangxiang Chu \\
  AMAP, Alibaba Group \\
  \texttt{chaoranxu\_yimo@163.com}
}
\begin{document}

\maketitle

\begin{abstract}
Vision-language models (VLMs) have achieved strong multimodal reasoning capabilities, but further improving them still relies heavily on large-scale human-constructed supervision for post-training. Such supervision is costly to obtain, especially for reasoning-intensive multimodal tasks where questions, answers, and feedback signals must be carefully designed.  This motivates self-evolving learning, where a model improves itself through a dual-role closed loop: a questioner autonomously poses questions and a solver learns to solve them. However, we observe that current VLM self-evolving methods still face three major challenges: coarse-grained role alternation delays the interaction between question generation and solver adaptation; generated questions can progressively degrade in quality; and question types may collapse toward a narrow distribution. These issues limit the efficiency and reliability of self-evolution. Thus, we propose \textbf{RISE}, a reliable self-evolving framework for vision-language models. RISE is built on three complementary designs: fine-grained role alternation, which shortens the feedback loop between the questioner and the solver to improve efficiency; a quality supervisor, which improves question validity and pseudo-label reliability; and skill-aware dynamic balancing, which mitigates mode collapse and maintains broad skill coverage during evolution.
Together, these components enable more reliable and effective self-evolution from unlabeled images. Experiments on two VLM backbones across seven benchmarks show that RISE consistently improves the base models, yielding broad and sustained gains. Our code is publicly available at https://github.com/AMAP-ML/RISE.
\end{abstract}

\section{Introduction}

Vision-language models (VLMs) have recently achieved substantial progress in multimodal understanding and reasoning~\cite{llava,qwen3,internvl3,llava15}. However, such progress has relied heavily on large-scale, high-quality human-annotated data. As increasingly powerful models~\cite{gpt,gemini,seed1,deepseek} consume much of the high-quality training resources available on the Internet, further scaling model capability by relying on externally curated supervision is becoming increasingly costly and less scalable~\cite{cambrian,survey,surveyvision,entropy}. This challenge is even more pronounced for VLMs: while large amounts of unlabeled images are readily available, high-quality visual reasoning supervision remains scarce, especially data that jointly couple images, questions, and answers in a reasoning-intensive manner~\cite{limitsofdata,limitsofvlm,limitsofvlm2}. This raises an increasingly important question: can VLMs rely more on self-generated learning signals from unlabeled images to achieve sustained capability improvement~\cite{deng2025self}?

In the large language model (LLM) domain, self-evolving frameworks such as R-Zero~\cite{rzero} and Absolute Zero~\cite{absolute} offer a promising route toward this goal~\cite{agent0,spice,guided}. These methods cast learning as a closed-loop interaction between two complementary roles: one role proposes tasks near the current capability boundary, while the other attempts to solve them, so that the model can autonomously construct training signals and improve itself without relying on human annotations~\cite{surveyselfevolutionlargelanguage}. Inspired by this paradigm, recent studies have begun to transfer this self-evolving paradigm to the vision-language setting. For example, works such as VisPlay~\cite{visplay} and MM-Zero~\cite{mmzero} extend this idea to visual inputs, enabling VLMs to  improve themselves by learning from the self-generated data. This migration from LLM self-evolution to VLM self-evolution provides a natural and appealing direction for unlabeled visual reasoning learning.

\begin{figure}[t]
	\centering
	\includegraphics[width=\textwidth]{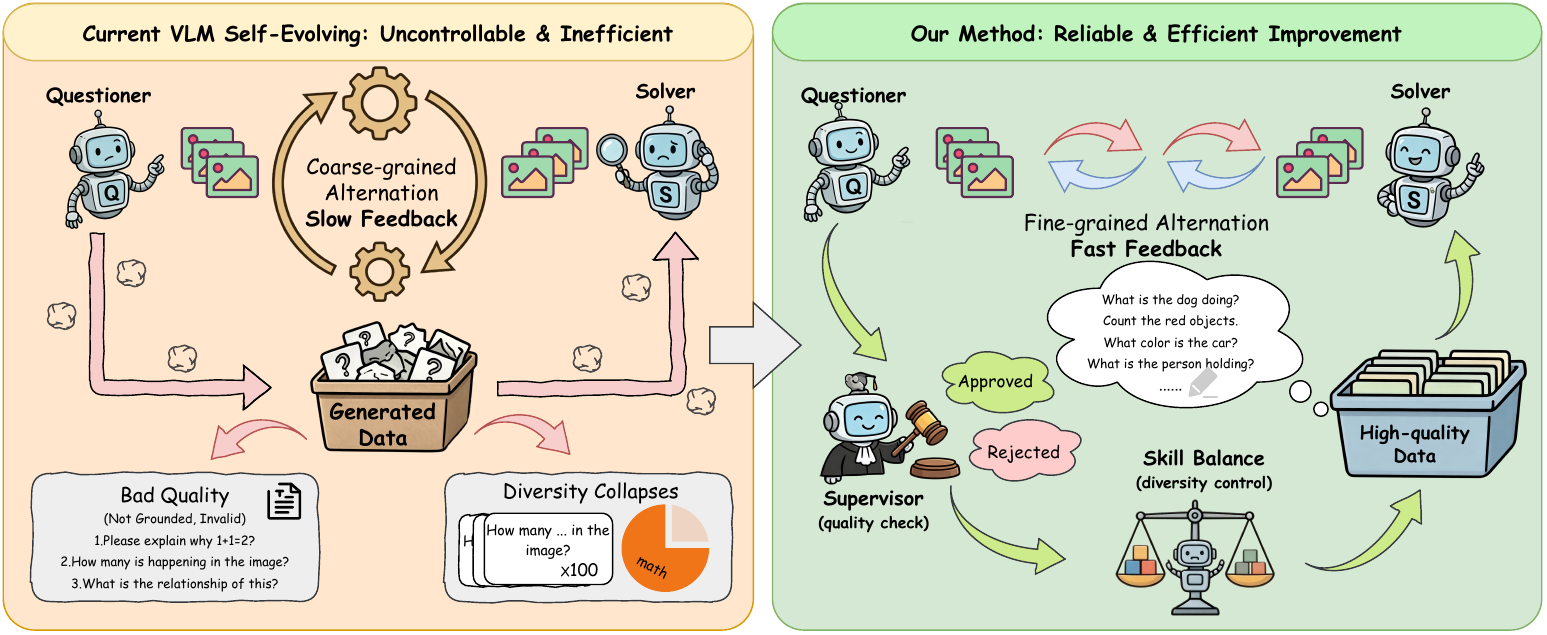}
	\caption{Comparison between current VLM self-evolving frameworks and our RISE framework.}
	\label{fig:teaser}
\end{figure}

However, directly transferring such dual-role self-evolving paradigms to VLMs is far from straightforward. As shown in Fig.~\ref{fig:teaser} (left), \textbf{firstly,} some existing methods~\cite{visplay,mmzero} typically adopt a coarse-grained role alternation scheme, where the questioner and the solver are optimized in separate long phases. This delayed interaction makes it difficult for question generation and solver adaptation to stay well coordinated, thereby reducing the overall efficiency of the self-evolving loop.
\textbf{Secondly,} question generation becomes substantially more challenging in the multimodal setting. Unlike in LLM self-evolution, where the model only needs to propose text-based tasks, a VLM must first understand the visual content and then formulate meaningful image-grounded questions. This added difficulty makes self-generated supervision much more prone to degeneration. In our experiments, we observe that without explicit control, self-evolution gradually suffers from both question quality degradation and diversity collapse. Specifically, the generated questions increasingly exhibit semantic errors, pseudo-difficult questions that are weakly grounded in the image, and even invalid questions. Meanwhile, the question distribution progressively collapses toward a few easy-to-generate categories, such as math and counting. Such degeneration weakens the usability and coverage of self-generated supervision, ultimately limiting sustained capability improvement. Therefore, effective self-evolving learning for VLMs requires not only more efficient role interaction, but also mechanisms that improve the reliability of self-generated supervision, especially in terms of question quality and diversity.

To address these issues (Fig.~\ref{fig:teaser} (right)), we first introduce \emph{fine-grained role alternation}, a more online-style training strategy that divides long role-switching phases into shorter update cycles and thereby shortens the feedback loop for role adaptation. On top of this, we further introduce a unified mechanism for question quality and diversity control. Specifically, we employ a \emph{quality supervisor} to constrain question validity and verify pseudo-label reliability, reducing the interference of low-quality questions and noisy samples. We also introduce \emph{skill-aware dynamic balancing} to dynamically regulate the question distribution and alleviate the progressive collapse of generated questions toward a narrow set of modes. In this way, fine-grained alternation improves the timeliness of role coordination, while quality and diversity control ensures the usability of self-generated supervision. Together, they support a more reliable and effective self-evolving process for VLMs.

Our contributions are as follows:
\begin{itemize}
	\item We identify the key limitations of existing VLM self-evolving methods, including feedback lag caused by coarse-grained role alternation, as well as the progressive degradation of question quality and collapse of diversity during self-evolution.
	\item We propose a set of reliability-oriented mechanisms for VLM self-evolution, including fine-grained role alternation to reduce feedback lag, a quality supervisor to improve question validity and pseudo-label reliability, and skill-aware dynamic balancing to mitigate diversity collapse and maintain broad supervision coverage.
	\item We present \textbf{RISE}, a self-evolving framework for vision-language models, which integrates the above designs into a unified training pipeline and enables more reliable and effective improvement from unlabeled images.
\end{itemize}

\section{Related Work}

\paragraph{Post-Training for Vision-Language Models.}
Recent post-training of vision-language models (VLMs) has evolved from supervised instruction tuning toward reinforcement learning-based optimization. Early works such as LLaVA mainly rely on visual instruction tuning to align visual encoders with language model backbones and improve multimodal instruction-following ability~\cite{llava,sftvisual,cotbal}. More recent R1-style and RLVR-based methods further show that reinforcement learning can enhance multimodal reasoning, visual understanding, and generalization when suitable reward signals are available~\cite{vlmr1,mmeureka,deng2025openvlthinker,sftorrl,chu2026gpg,harder,ranking,wang2026visually}. However, most existing post-training methods still depend on human-curated instruction data, annotated reasoning traces, or externally constructed verifiable tasks, which are costly to scale for reasoning-intensive visual learning~\cite{r1-onevision,visionr1}. This motivates self-evolving VLMs, where models exploit unlabeled images and self-generated supervision to construct their own training signals with minimal human annotation.

\paragraph{Self-Evolving Learning.} Recent work has shown that large language models can improve themselves with little or no human supervision by exploiting self-generated learning signals, such as pseudo-labels, self-consistency, and verifiable rewards~\cite{boostrap,huang2023large,wang2023self,gulcehre2023reinforced,singh2023beyond,skillclaw,coevolve}. Early efforts such as \emph{TTRL}~\cite{zuo2025ttrl} explored how models can learn from self-generated weak supervision without relying on explicit ground-truth labels, while more recent dual-role frameworks, including \emph{R-Zero}~\cite{rzero} and \emph{Absolute Zero}~\cite{absolute}, further cast self-evolution as a closed loop between complementary roles that generate and solve tasks. Inspired by these advances in LLMs, recent studies have extended this paradigm to vision-language models, including \emph{VisPlay}~\cite{visplay}, \emph{Vision-Zero}~\cite{vision}, \emph{EvoLMM}~\cite{evolmm}, and \emph{MM-Zero}~\cite{mmzero}, showing that multimodal models can also benefit from self-play supervision under unlabeled or weakly supervised settings. However, VLM self-evolution is substantially more fragile, since question generation must remain visually grounded, making the intermediate supervision more prone to quality degradation and diversity collapse. Therefore, rather than further establishing the self-evolving pipeline itself, our work focuses on improving its reliability through better interaction efficiency, question quality control, and skill diversity maintenance.
\section{Preliminaries}

\subsection{Group Relative Policy Optimization}

Group Relative Policy Optimization (GRPO) is a reinforcement learning~\cite{rl} method that updates a policy by comparing multiple responses generated for the same input, without relying on a separately learned value model. Its key idea is to normalize rewards within a response group and use the resulting relative advantages for policy optimization.

Given an input prompt $p$, the current policy $\pi_{\theta_{\mathrm{old}}}$ generates a group of $G$ responses $\{x_1,\ldots,x_G\}$. Each response $x_i$ is assigned a scalar reward $r_i$. The relative advantage of response $x_i$ is computed as
\begin{equation}
	\hat{A}_i=
	\frac{r_i-\mathrm{mean}(r_1,\ldots,r_G)}
	{\mathrm{std}(r_1,\ldots,r_G)+\epsilon_{\mathrm{norm}}},
\end{equation}
where $\epsilon_{\mathrm{norm}}$ is a small constant for numerical stability. The policy is then updated with a clipped objective:
\begin{equation}
	\mathcal{L}_{\mathrm{GRPO}}(\theta)
	=
	-\frac{1}{G}\sum_{i=1}^{G}
	\min\left(
	\frac{\pi_\theta(x_i)}{\pi_{\theta_{\mathrm{old}}}(x_i)}\hat{A}_i,\,
	\mathrm{clip}\!\left(
	\frac{\pi_\theta(x_i)}{\pi_{\theta_{\mathrm{old}}}(x_i)},
	1-\epsilon,1+\epsilon
	\right)\hat{A}_i
	\right).
\end{equation}

Such an optimization strategy is well suited to our VLM self-evolving framework, where model-generated answers are directly verifiable and therefore naturally amenable to RLVR.
\subsection{Dual-Role Self-Evolving Pipeline}

Recent self-evolving frameworks for LLMs and VLMs, such as \emph{R-Zero} and \emph{VisPlay}, typically follow a dual-role iterative pipeline. 
Under this paradigm, the questioner is first optimized according to feedback from the current solver so as to generate questions near the solver's capability boundary. Then, pseudo-labeled training data are constructed from the updated questioner and the majority-voted solver responses, and the solver is subsequently optimized on these data. A conventional dual-role self-evolving pipeline can be written as
\begin{equation}
	\pi_q^{(t+1)}=\mathrm{Update}_q^{B}\!\big(\pi_q^{(t)};\pi_s^{(t)}\big),\;
	\mathcal{D}^{(t)}=\mathrm{Construct}\!\big(\pi_q^{(t+1)},\pi_s^{(t)}\big),\;
	\pi_s^{(t+1)}=\mathrm{Update}_s^{B}\!\big(\pi_s^{(t)};\mathcal{D}^{(t)}\big).
	\label{eq:coarse_alt_pipeline}
\end{equation}
where $\pi_q^{(t)}$ and $\pi_s^{(t)}$ denote the questioner and the solver at self-evolving round $t$, respectively; $\mathrm{Update}_q^{B}$ and $\mathrm{Update}_s^{B}$ denote applying $B$ successive optimization steps to the questioner and the solver, respectively; and $\mathcal{D}^{(t)}=\{(x,q,\hat{y})\}$ denotes the pseudo-labeled training data constructed at self-evolving round $t$ for solver training.

In this way, question generation and question solving together form a closed self-improvement loop, which serves as the common foundation of existing dual-role self-evolving frameworks.
\begin{figure}[t]
	\centering
	\includegraphics[width=\textwidth]{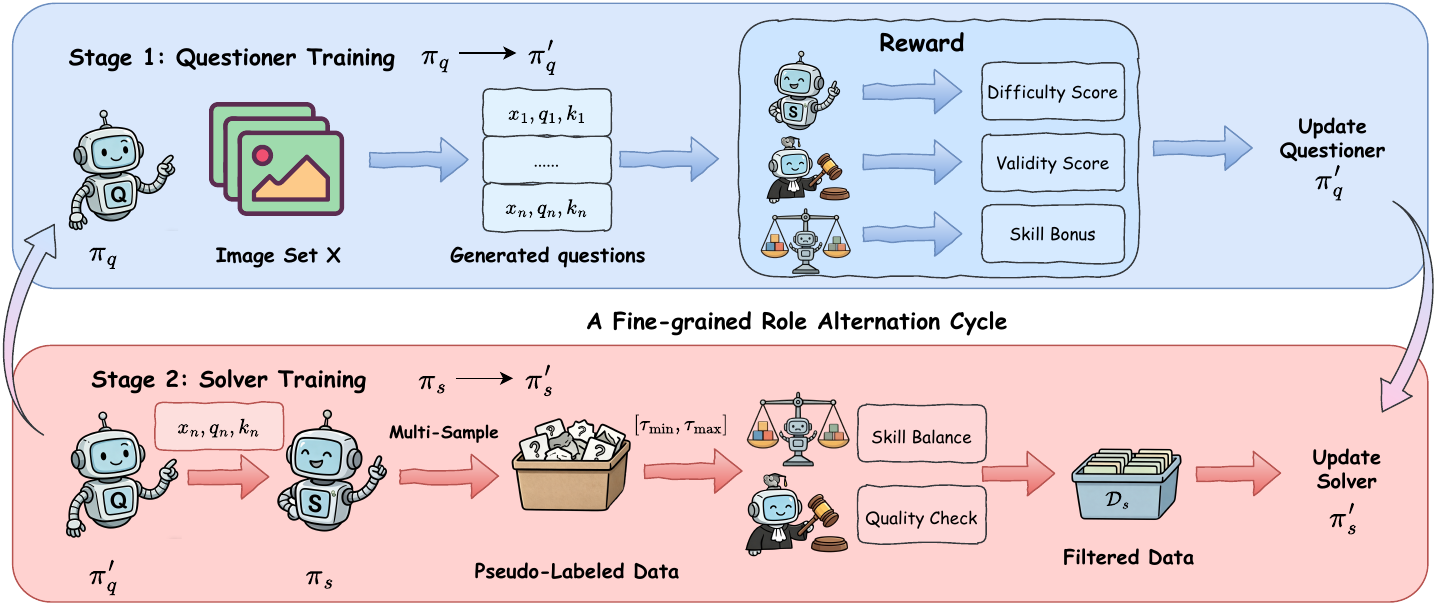}
	\caption{Overview of \textbf{RISE}, a dual-role self-evolving framework for VLMs that alternates questioner and solver training with quality supervision and skill balancing.
	}
	\label{fig:method}
\end{figure}

\section{Method}

\subsection{Overview}

As shown in Fig.~\ref{fig:method}, given an unlabeled image set $\mathcal{X}=\{x_i\}_{i=1}^{N}$, we use a single vision-language model to play two functional roles: a \emph{questioner} $\pi_q$ and a \emph{solver} $\pi_s$. Our framework is built under a \emph{fine-grained role alternation} training strategy, which enables more timely feedback exchange between the two roles during self-evolution. Under this strategy,  we  first train the questioner to generate informative questions for $\mathcal{X}$, using reward signals derived from the solver's response quality and the supervisor feedback. Second, based on the questions generated by the current questioner and the corresponding sampled responses from the solver, we construct and filter pseudo-labeled training data. Third, we train the solver on the filtered pseudo-labeled data to improve visual reasoning capability. The framework is further equipped with two quality and diversity control mechanisms: a quality supervisor to ensure question quality and pseudo-label reliability, and a skill-aware dynamic balancing mechanism to maintain diverse skill coverage during self-evolution.

\subsection{Fine-Grained Role Alternation}

As formalized in Eq.~(\ref{eq:coarse_alt_pipeline}), conventional dual-role self-evolution typically relies on \emph{coarse-grained role alternation}: within one alternation round, the questioner keeps adapting to the old solver policy $\pi_s^{(t)}$ for an entire phase of $B$ update steps, while the solver is later trained for another phase of $B$ update steps on data induced by this stale interaction. As a result, the two roles are synchronized only once per round, which slows down capability alignment between them.

To mitigate this lag, we adopt a \emph{fine-grained role alternation} strategy. Instead of switching roles only once after $B$ update steps, we split the same overall update budget into $n$ shorter phases of $b$ steps.
We define one short alternation cycle of length $b$ as
\begin{equation}
	\pi_q'=\mathrm{Update}_q^{b}\!\big(\pi_q;\pi_s\big),\qquad
	\mathcal{D}'=\mathrm{Construct}\!\big(\pi_q',\pi_s\big),\qquad
	\pi_s'=\mathrm{Update}_s^{b}\!\big(\pi_s;\mathcal{D}'\big).
\end{equation}
Then our fine-grained role alternation can be summarized as
\begin{equation}
	\big(\pi_q^{(t+1)},\pi_s^{(t+1)}\big)=\mathrm{Alt}_{b}^{\,n}\!\big(\pi_q^{(t)},\pi_s^{(t)}\big),\qquad B=nb,
\end{equation}
where $\mathrm{Alt}_{b}$ denotes one short alternation cycle. In this way, the feedback loop between question generation and solver adaptation is substantially shortened: the questioner can receive feedback from a more up-to-date solver after every short phase, and the solver can also be updated using data produced by a more up-to-date questioner, while keeping the same overall update budget. Consequently, the two roles can complete capability alignment more quickly, making the overall self-evolving process more efficient and allowing the model to reach a stronger performance regime earlier.

\subsection{Quality and Diversity Control}

\subsubsection{Quality Supervisor}
If the questioner is optimized only for generating questions near the solver's capability boundary, it may exploit this objective by producing pseudo-difficult but meaningless questions, such as weakly grounded, semantically ill-posed, skill-mismatched, or visually unanswerable ones. To avoid this issue, we introduce a supervisor to constrain question quality and verify pseudo-label reliability.

Instead of training an additional judge model, we let the supervisor share parameters with the current solver, i.e., $\pi_{\mathrm{sup}} \equiv \pi_s$. This shared-parameter design introduces no extra model cost and also improves internal consistency: a model that answers questions should also be able to assess whether those questions and the generated answers are acceptable. Meanwhile, we do not let the questioner itself act as the supervisor, which helps reduce reward hacking by avoiding self-judging.

Given an image $x$, the questioner generates a structured output $(k,q)\sim \pi_q(\cdot\mid x)$, where $k$ denotes the declared skill and $q$ is the question text.
The supervisor serves two purposes in our framework. First, it participates in questioner training by judging whether a generated question is valid. We use a binary signal $v(x,q,k)\in\{0,1\}$, which equals $1$ only if the question is image-grounded, semantically well-formed, answerable from the image alone, and consistent with its declared skill. This signal is directly incorporated into the questioner reward so that the model cannot obtain high reward merely by confusing the solver.

Second, it is used during pseudo-label data construction to verify whether the candidate answer is acceptable. We define another binary signal $u(x,q,\hat{y})\in\{0,1\}$  to verify the correctness of the majority-voted answer $\hat{y}$ produced by the solver.

In practice, both $v(x,q,k)$ and $u(x,q,\hat{y})$ are obtained by prompt-based judgment using the supervisor.

\subsubsection{Skill-Aware Dynamic Balancing}

During self-evolution, the questioner tends to gradually favor a few skills that yield higher reward more easily, such as counting or math. Such mode collapse makes the generated questions increasingly narrow and limits the range of reasoning skills the model can acquire.

To address this issue, we introduce a skill-aware dynamic balancing mechanism. The key idea is to maintain the recent frequency of each skill during training and provide additional encouragement to under-represented skills, thereby promoting more diverse question generation.

We define the skill set $\mathcal{K}$ as comprising six categories: coarse perception, fine-grained perception, instance reasoning, logical reasoning, math \& counting, and science \& technology. For each skill $k\in\mathcal{K}$, let $n_k$ denote the recent number of generated questions whose declared skill is $k$ before the current alternation cycle. We compute the average skill frequency as
\begin{equation}
	\bar n=\frac{1}{|\mathcal{K}|}\sum_{k'\in\mathcal{K}} n_{k'}.
\end{equation}
For a question with declared skill $k$, we define the skill bonus as
\begin{equation}
	b_{\mathrm{skill}}(k)=
	\max\left(
	\frac{\bar n-n_k}{\bar n},
	0
	\right).
\end{equation}
In this way, only relatively under-represented skills receive additional reward, while dominant skills receive no extra encouragement. Combined with fine-grained role alternation, it also enables more timely adjustment of the question skill distribution during self-evolution. 

We further apply skill-stratified sampling when constructing the final pseudo-labeled data, by drawing approximately equal-sized subsets from the six skill categories, so that skill balancing affects both question generation and solver training.

\subsection{Training Objectives}

\subsubsection{Questioner Objective}

The questioner is encouraged to generate questions near the solver's capability boundary. To assess this, we first let the solver sample $M$ responses:
\begin{equation}
	y_j\sim \pi_s(\cdot \mid x,q),\qquad j=1,\ldots,M.
\end{equation}
Based on these responses, we compute the self-consistency score
\begin{equation}
	c(x,q)=\frac{1}{M}\max_a \sum_{j=1}^{M}\mathbf{1}[y_j=a].
\end{equation}
A larger $c(x,q)$ indicates that the solver responses are more stable for the question $(x,q)$. We use this score to construct the difficulty score~\cite{sun2025improving}
\begin{equation}
	d(x,q)=\min\big(c(x,q),\,1-c(x,q)\big),
\end{equation}
which favors questions near the solver's current capability boundary rather than overly easy or highly unstable ones.

The questioner reward is defined as
\begin{equation}
	r_q(x,q,k)=
	\left\{
	\begin{array}{ll}
		-1, & \mbox{invalid format}, \\[4pt]
		d(x,q)+\lambda_v\,v(x,q,k)+\lambda_s\,b_{\mathrm{skill}}(k), & \mbox{valid format}.
	\end{array}
	\right.
	\label{eq:rq}
\end{equation}
which encourages the questioner to generate questions that are moderately difficult, semantically valid, and diverse in skill coverage.
\subsubsection{Solver Objective}

Based on the questions generated by the questioner trained in the current alternation round and the corresponding sampled solver responses, we construct solver training data. For each candidate question, the pseudo label is obtained by majority voting over the sampled solver outputs:
\begin{equation}
	\hat{y}=\arg\max_a \sum_{j=1}^{M}\mathbf{1}[y_j=a].
\end{equation}
We then reuse the same self-consistency score $c(x,q)$ to estimate sample confidence, and pre-filter candidate samples by
\begin{equation}
	c(x,q)\in[\tau_{\min},\tau_{\max}],
\end{equation}
where overly easy questions usually provide limited learning value, while highly unstable ones are often noisy. After this confidence-based pre-filtering, we further retain only samples satisfying
\begin{equation}
	v(x,q,k)=1,\qquad u(x,q,\hat{y})=1.
\end{equation}
We then apply skill-balanced sampling over the remaining samples to form the final pseudo-labeled training set, which is defined as
\begin{equation}
	\mathcal{D}_s=\left\{(x,q,\hat{y},k)\right\}.
\end{equation}

The solver is trained on the filtered pseudo-labeled training set $\mathcal{D}_s$. Given a sample $(x,q,\hat y,k)$ and solver output $\tilde y$, the solver reward is defined as
\begin{equation}
	r_s(\tilde y,\hat y)=
	\left\{
	\begin{array}{ll}
		-1, & \mbox{invalid format}, \\[4pt]
		\mathbf{1}[\mathrm{extract}(\tilde y)=\hat y], & \mbox{valid format}.
	\end{array}
	\right.
\end{equation}

\section{Experiments}

\subsection{Experimental Setup}

\paragraph{Models and Image Set.}
To validate the effectiveness of the proposed self-evolving framework, we conduct experiments on two vision-language models of different scales and generations, Qwen2.5-VL-7B~\cite{qwen25vltechnicalreport} and Qwen3-VL-8B~\cite{qwen3}. In our experiments, we initialize the questioner and the solver from the same base model, and train them under the same self-evolving pipeline. For self-evolving training, we use the unlabeled image collection from Vision-47K~\cite{video,vision47}, which  contains about 47K images and covers diverse visual scenes and task sources, providing a diverse image pool.

\paragraph{Training Details.}
Following the setting in prior coarse-grained dual-role self-evolving frameworks, we use a total update budget of $B=20$ as the reference within one alternation round. For fair comparison, our fine-grained role alternation keeps the same total budget and sets $n=4$ and $b=5$, such that $B = nb = 20$. In each update step, we sample 256 unlabeled images for self-evolving training. For each question generated by the questioner, the solver samples $M=10$ responses for difficulty estimation and pseudo-label construction. For pseudo-label filtering, we set the self-consistency threshold range to $[\tau_{\min}, \tau_{\max}] = [0.3, 0.8]$. In the questioner reward in Eq.~(\ref{eq:rq}), we set the weights of the validity term and the skill bonus term to $\lambda_v = 0.2$ and $\lambda_s = 0.2$, respectively.

\paragraph{Evaluation Benchmarks and Protocol.}
We evaluate the evolved models on MMMU~\cite{mmmu}, MMVet~\cite{mmvet}, RealWorldQA~\cite{realworldqa}, ChartQA~\cite{chartqa}, MathVista~\cite{mathvista}, MathVerse~\cite{mathverse}, and MathVision~\cite{mathvision}. These benchmarks jointly cover a broad range of capabilities, including general visual understanding, chart understanding, and multimodal mathematical reasoning, thereby providing a comprehensive testbed for assessing whether self-evolution improves both general-purpose and reasoning-intensive visual capabilities.
We compare the models before and after self-evolution under the same inference setting on all evaluation benchmarks. Following an LLM-as-a-Judge protocol~\cite{judge0,li2024llms}, we use Qwen3-Max to assess answer correctness, as it better handles semantically equivalent answers than exact string matching. In addition, we compute an overall score as the sample-size weighted average over all benchmarks. More details are provided in Appendix~\ref{app:training_details}.

\subsection{Main Results}
\begin{table*}[t]
	\centering
	\footnotesize
	\setlength{\tabcolsep}{4.2pt}
	\renewcommand{\arraystretch}{0.9}
	\begin{tabular}{lcccccccc}
		\toprule
		\textbf{Model Name} & \textbf{MMMU} & \textbf{MMVet} & \textbf{RealWQA} & \textbf{ChartQA} & \textbf{MathVerse} & \textbf{MathVision} & \textbf{MathVista} & \textbf{AVG} \\
		\midrule
		\multicolumn{9}{c}{\textbf{Qwen2.5-VL-7B-Instruct}} \\
		\midrule
		Base Model   & 49.71 & 51.83 & 57.12 & 79.48 & 40.48 & 23.21 & 58.70 & 48.17 \\
		RISE-Step20 & 54.03 & \textbf{57.80} & 57.52 & 81.68 & 43.65 & 27.36 & 60.40 & 51.20 \\
		RISE-Step40 & 54.03 & 56.88 & 58.30 & 81.84 & 44.19 & \textbf{28.73} & 60.10 & 51.74 \\
		RISE-Step60 & \textbf{55.31} & \textbf{57.80} & \textbf{60.39} & \textbf{82.08} & \textbf{44.95} & 28.32 & \textbf{61.10} & \textbf{52.26} \\
		\midrule
		\multicolumn{9}{c}{\textbf{Qwen3-VL-8B-Instruct}} \\
		\midrule
		Base Model    & 55.31 & 60.55 & 67.45 & 81.44 & 46.75 & 27.32 & 65.80 & 53.38 \\
		RISE-Step20 & 55.54 & 61.93 & 68.50 & 82.28 & 48.65 & 31.28 & 65.60 & 55.17 \\
		RISE-Step40 & 58.11 & 62.39 & \textbf{70.85} & 82.24 & 49.06 & 32.80 & 67.70 & 56.15 \\
		RISE-Step60 & \textbf{59.63} & \textbf{62.84} & 70.33 & \textbf{83.64} & \textbf{49.11} & \textbf{36.10} & \textbf{68.70} & \textbf{57.37} \\
		\bottomrule
	\end{tabular}
	\caption{Main results of RISE on two base VLMs under different self-evolving steps. We report accuracy on seven benchmarks and the sample-size weighted average (AVG).}
	\label{tab:main_results}
\end{table*}
\begin{table*}[t]
	\centering
	\footnotesize
	\setlength{\tabcolsep}{4.4pt}
	\renewcommand{\arraystretch}{0.9}
	\begin{tabular}{lccccccc}
		\toprule
		\textbf{Method} & \textbf{MMMU} & \textbf{MMVet} & \textbf{RealWQA} & \textbf{ChartQA} & \textbf{MathVerse} & \textbf{MathVision} & \textbf{MathVista} \\
		\midrule
		\multicolumn{8}{c}{\textbf{Qwen2.5-VL-7B-Instruct}} \\
		\midrule
		\textbf{RISE}                         & \textbf{+5.60} & \textbf{+5.97} & \textbf{+3.27} & +2.60 & \textbf{+4.47} & \textbf{+5.11} & +2.40 \\
		Vision-Zero                 & +1.60          & --             & +0.40          & +1.10 & +3.00          & +3.50          & \textbf{+3.90} \\
		EvoLMM            & +0.90          & --             & --             & \textbf{+2.70} & +1.10          & +0.90          & +2.06 \\
		VisPlay$^\dagger$                     & +3.38          & +4.59          & +1.96          & +1.96 & +2.62          & -0.40          & +1.80 \\
		\midrule
		\multicolumn{8}{c}{\textbf{Qwen3-VL-8B-Instruct}} \\
		\midrule
		\textbf{RISE}                         & \textbf{+4.32} & \textbf{+2.29}          & \textbf{+2.88} & \textbf{+2.20} & +2.36          & \textbf{+8.78} & \textbf{+2.90} \\
		MM-Zero                              & +2.50          & +0.90 & +0.79          & +2.00          & \textbf{+4.50} & +8.10          & -0.50 \\
		\bottomrule
	\end{tabular}
	\caption{Comparison with prior self-evolving VLM methods using gain over the corresponding base model. $^\dagger$ denotes our reproduced result, while  other  results are  taken from the corresponding papers.}
	\label{tab:gain_comparison}
\end{table*}

\paragraph{Self-Evolving Results.}
Table~\ref{tab:main_results} summarizes the self-evolving results of RISE on Qwen2.5-VL-7B-Instruct and Qwen3-VL-8B-Instruct. RISE consistently improves both backbones across self-evolving steps. On Qwen2.5-VL-7B-Instruct, the weighted average score increases from 48.17 to 51.20, 51.74, and 52.26 after 20, 40, and 60 steps, respectively, yielding a total gain of +4.09. On Qwen3-VL-8B-Instruct, the average score improves from 53.38 to 55.17, 56.15, and 57.37, corresponding to a total gain of +3.99 after 60 steps. These gains are broad across evaluation domains, including general visual understanding, chart understanding, and multimodal mathematical reasoning, with particularly strong improvements on reasoning-intensive benchmarks such as MathVerse and MathVision. We attribute these stable improvements to the explicit control of question quality and skill distribution during self-evolution, which provides more reliable and learning-effective supervision throughout the intermediate training process.
We also observe that, on most benchmarks, the marginal gains become smaller in later stages, suggesting a gradual convergence effect. From a methodological perspective, the solver is updated by using pseudo labels constructed from majority-voted responses, i.e., a maj@$n$ policy, to improve the current policy. While this form of supervision is effective in correcting evident errors at earlier stages, its incremental benefit naturally diminishes as the solver becomes stronger. As a result, self-evolution continues to yield positive gains in later stages, but the improvement typically becomes more moderate over time.

\paragraph{Comparison with Prior Self-Evolving Methods.}
We further provide a reference comparison between RISE and prior self-evolving VLM methods, including image-driven self-evolving frameworks such as VisPlay (CVPR'26), EvoLMM (CVPR'26 Findings), and Vision-Zero (ICLR'26), as well as MM-Zero, which follows a zero-data setting without relying on external real images. Considering that these methods differ in base model implementations, training configurations, and evaluation settings, we report the improvement margins in Table~\ref{tab:gain_comparison} over their corresponding base models as a reference for comparing their practical gains. From the results, RISE shows relatively strong improvement margins across different backbones and multiple benchmarks compared with prior methods.
\paragraph{Analysis of the Quality and Diversity.}
\begin{figure}[t]
	\centering
	\includegraphics[width=\textwidth]{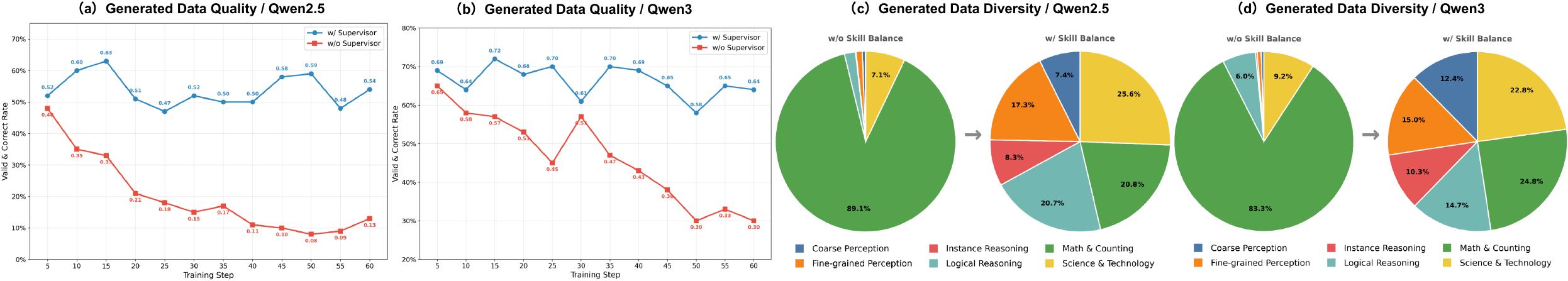}
	\caption{Statistics of question quality and diversity during self-evolution.}
	\label{fig:question}
\end{figure}
We conduct a statistical analysis of the intermediate VQA data generated during self-evolution, as shown in Fig.~\ref{fig:question}, to verify the degeneration phenomena and the effectiveness of our method.
For \textbf{question quality}, we evaluate the validity and correctness of generated questions. Without a supervisor, the valid-and-correct rate declines steadily and drops below 10\% in later stages, indicating the accumulation of low-quality questions. In contrast, introducing the supervisor keeps question quality consistently high, showing that it effectively prevents degradation and ensures reliable intermediate supervision.
For \textbf{question diversity}, the skill distribution shows that without skill balance, generated questions collapse toward a few high-reward skills. This is mainly because the questioner reward includes the difficulty score, which gradually drives the model to favor skill types that are easier to obtain high difficulty rewards from, thereby narrowing the coverage of training signals. With skill-aware dynamic balancing, the skill distribution becomes more balanced, effectively alleviating mode collapse. These two phenomena can be consistently observed during the self-evolution of both base models. Additional qualitative examples and statistics are provided in Appendix~\ref{app:qualitative}, and the reliability of the supervisor judgment is further analyzed in Appendix~\ref{app:supervisor_pr}.

\subsection{Ablation Study on RISE Components}

\begin{table*}[t]
	\centering
	\scriptsize
	\setlength{\tabcolsep}{2.6pt}
	\renewcommand{\arraystretch}{0.92}
	\begin{tabular}{llcccccccc}
		\toprule
		\textbf{Step} & \textbf{Variant} & \textbf{MMMU} & \textbf{MMVet} & \textbf{RealWQA} & \textbf{ChartQA} & \textbf{MathVerse} & \textbf{MathVision} & \textbf{MathVista} & \textbf{AVG} \\
		\midrule
		-- & Base Model & 49.71 & 51.83 & 57.12 & 79.48 & 40.48 & 23.21 & 58.70 & 48.17 \\
		\midrule
		\multirow{5}{*}{\textbf{Step 20}} 
		& Fine-grained w/o Sup.   & 51.58 & \textbf{59.63} & 57.65 & \textbf{81.68} & 42.89 & 26.80 & 57.30 & 50.43 \\
		& Coarse-grained w/o Sup. & 52.51 & 56.42 & 58.17 & 80.76 & 41.83 & 24.66 & 58.90 & 49.58 \\
		& Fine-grained + Sup.     & 53.68 & 57.80 & 58.04 & 81.08 & 42.69 & 27.03 & 59.80 & 50.64 \\
		& Coarse-grained + Sup.   & 51.23 & 58.26 & \textbf{58.56} & 81.56 & 41.83 & 24.51 & \textbf{60.40} & 49.81 \\
		& \textbf{RISE}           & \textbf{54.03} & 57.80 & 57.52 & \textbf{81.68} & \textbf{43.65} & \textbf{27.36} & \textbf{60.40} & \textbf{51.20} \\
		\midrule
		\multirow{5}{*}{\textbf{Step 40}} 
		& Fine-grained w/o Sup.   & 52.51 & 56.42 & 57.78 & 81.72 & 42.39 & 26.40 & 60.00 & 50.42 \\
		& Coarse-grained w/o Sup. & 52.74 & 56.42 & 59.35 & 81.04 & 43.48 & 24.99 & 58.60 & 50.32 \\
		& Fine-grained + Sup.     & 52.86 & \textbf{60.09} & 58.82 & 81.64 & \textbf{45.15} & 27.21 & 59.90 & 51.65 \\
		& Coarse-grained + Sup.   & 51.81 & \textbf{60.09} & \textbf{59.87} & 81.44 & 43.93 & 24.92 & \textbf{61.30} & 50.80 \\
		& \textbf{RISE}           & \textbf{54.03} & 56.88 & 58.30 & \textbf{81.84} & 44.19 & \textbf{28.73} & 60.10 & \textbf{51.74} \\
		\midrule
		\multirow{5}{*}{\textbf{Step 60}} 
		& Fine-grained w/o Sup.   & 41.19 & 51.38 & 55.56 & 80.00 & 33.83 & 21.62 & 56.60 & 44.85 \\
		& Coarse-grained w/o Sup. & 53.09 & 56.42 & 59.08 & 81.44 & 43.10 & 22.81 & 60.50 & 49.96 \\
		& Fine-grained + Sup.     & 53.56 & 56.42 & 57.25 & 82.00 & \textbf{45.18} & 27.51 & 60.20 & 51.71 \\
		& Coarse-grained + Sup.   & 54.38 & 57.34 & 58.82 & 81.48 & 44.67 & 23.99 & 60.00 & 50.80 \\
		& \textbf{RISE}           & \textbf{55.31} & \textbf{57.80} & \textbf{60.39} & \textbf{82.08} & 44.95 & \textbf{28.32} & \textbf{61.10} & \textbf{52.26} \\
		\bottomrule
	\end{tabular}
	\caption{Ablation study of RISE on Qwen2.5-VL-7B-Instruct. ``Fine-grained'' and ``Coarse-grained'' denote the role alternation strategy, and ``Sup.'' denotes the quality supervisor. All non-RISE variants remove the skill balancing module, while RISE denotes the full model.}
	\label{tab:ablation}
\end{table*}
Table~\ref{tab:ablation} reports the ablation results of RISE on Qwen2.5-VL-7B-Instruct. We analyze the effects of the three core components in RISE, and summarize four main observations. Additional ablation study and analysis  are provided in Appendix~\ref{app:additional_experiments}.

\textbf{(1) Fine-grained role alternation improves self-evolving efficiency.}
Regardless of whether the quality supervisor is used, fine-grained role alternation generally reaches a stronger performance regime with fewer update steps than its coarse-grained counterpart. For example, without the supervisor, \emph{Fine-grained w/o Sup.} already achieves an AVG of 50.43 at step 20, which is higher than \emph{Coarse-grained w/o Sup.} at step 40 (50.32) and step 60 (49.96). A similar trend is also observed when the supervisor is enabled. These results show that shortening the feedback loop between the questioner and the solver makes self-evolution more efficient. We further analyze the effect of different alternation granularities in Appendix~\ref{app:alternation_granularity}.

\textbf{(2) Faster alternation alone can aggravate quality degradation.}
Although fine-grained alternation is more efficient, it is not sufficient by itself to ensure stable evolution. Without the quality supervisor, the AVG of \emph{Fine-grained w/o Sup.} drops from 50.43  to 50.42, and then sharply to 44.85 at step 60. This suggests that when role interaction becomes faster but question quality is not explicitly constrained, the questioner is more likely to exploit the reward and drift toward low-quality or reward-hacked questions, which eventually harms pseudo-label quality and solver training.

\textbf{(3) The supervisor enables fast yet stable evolution.}
Comparing \emph{Fine-grained + Sup.} with \emph{Fine-grained w/o Sup.} shows that the  supervisor effectively stabilizes the fine-grained self-evolving process. With the supervisor, the AVG remains strong from 50.64 to 51.71, avoiding the severe collapse observed in the unsupervised fine-grained setting. Similar benefits can also be observed in the coarse-grained setting.  This result is consistent with our design motivation: the supervisor helps enforce question validity and pseudo-label reliability, thereby allowing fine-grained role alternation to retain its efficiency advantage without sacrificing training stability.

\textbf{(4) Skill balancing further improves diversity and final performance.}
On top of \emph{Fine-grained + Sup.} setting, adding skill balancing gives the full RISE model and yields the best overall performance. RISE brings further gains across steps, and these gains are broad across benchmarks rather than coming from a single dataset. This indicates that skill balancing helps prevent question generation from collapsing to a narrow set of easy-reward skills and supports more balanced capability growth.

\section{Conclusion}
In this paper, we present \textbf{RISE}, a reliable self-evolving framework for vision-language models under unlabeled images. By identifying the key limitations of existing methods, including delayed role interaction, question quality degradation, and diversity collapse, we develop a unified solution that combines fine-grained role alternation, a quality supervisor, and skill-aware dynamic balancing. These components work together to improve interaction efficiency, ensure the reliability of self-generated supervision, and maintain diverse skill coverage throughout the evolution process. Extensive experiments show that RISE consistently brings stable and sustained performance gains across different model scales and multiple benchmarks. Overall, our results suggest that reliable interaction and controlled supervision are crucial for effective and reliable VLM self-evolution.

{
% \small
\bibliographystyle{plainnat}
\bibliography{references}
}
\clearpage
\appendix
\section*{Appendix}
\section{Detailed Training and Implementation Details}
\label{app:training_details}

\subsection{Hyperparameter Settings}
\label{app:hyperparameters}

We use the same training hyperparameters for the questioner and the solver unless otherwise specified. In our fine-grained role alternation strategy, each short alternation cycle contains $b=5$ update steps. The whole self-evolving process contains 12 fine-grained alternation cycles. Therefore, the full training process contains 60 update steps in total, and we report the intermediate checkpoints after 20, 40, and 60 update steps.

For GRPO optimization, we use 8 rollouts per training prompt. During question evaluation and pseudo-label construction, the solver samples $M=10$ responses for each generated question, which are used to estimate self-consistency, compute the difficulty score, and construct majority-voted pseudo-labels. Following the main paper, the reward weights for the validity score and the skill bonus are both set to $\lambda_v=0.2$ and $\lambda_s=0.2$. The batch size is set to 256, which means that training for 20 update steps requires generating approximately 5K VQA samples. During rollout generation, we use a temperature of 1.0 and a top-$p$ value of 0.99.

For benchmark evaluation, we use a unified LLM-as-a-Judge protocol to assess answer correctness. This is because model responses to multimodal benchmarks can have diverse surface forms, and several benchmarks include open-ended VQA questions whose correct answers cannot be reliably evaluated by exact string matching alone. Judge-based evaluation is therefore more suitable for recognizing semantically equivalent answers expressed in different forms.
For each test sample, we generate one response using greedy decoding, and then use the judge model to determine whether the extracted answer matches the ground-truth answer.

All experiments are conducted on 8 NVIDIA H20 GPUs. The complete self-evolving training process takes approximately 48 hours.

\subsection{Prompt Templates}
\label{app:prompt_templates}

We use four types of prompts in RISE: question generation, solver answering, question validity judgment, and answer verification. For readability, we omit the model-specific chat template tokens and denote the input image as \texttt{[IMAGE]}. The placeholders \texttt{\{QUESTION\}}, \texttt{\{DECLARED\_SKILL\}}, \texttt{\{SKILL\_CONTEXT\}}, and \texttt{\{CANDIDATE\_ANSWER\}} are dynamically filled during training.

\paragraph{Questioner prompt.}
The questioner is instructed to generate one challenging visual reasoning question from the given image. It must declare one skill category and one question type, and output the result in a structured format.

\begin{verbatim}
	System:
	You are an intelligent Question Generator. Given an image, generate exactly
	one difficult visual reasoning question that is directly grounded in the image.
	
	The question must:
	1. require visual analysis or reasoning rather than simple description;
	2. belong to exactly one skill category:
	coarse perception, fine-grained perception, instance reasoning,
	logical reasoning, math & counting, or science & technology;
	3. belong to exactly one question type:
	multiple choice, numerical, or regression;
	4. have a short, unique, and verifiable answer.
	
	Output strictly in the following format:
	<skill>...</skill>
	<type>...</type>
	<question>...</question>
	
	User:
	[IMAGE]
	Generate one new challenging reasoning question based on this image.
\end{verbatim}

\paragraph{Solver prompt.}
The solver is prompted to answer the generated question based on the image. To make answer extraction reliable, the final answer is required to appear inside \texttt{\textbackslash boxed\{\}}.

\begin{verbatim}
	System:
	You are a helpful visual reasoning assistant.
	
	User:
	[IMAGE]
	Please reason step by step carefully based on the image for the following
	question:
	{QUESTION}
	
	After completing your reasoning, output the final clean and concise answer
	strictly inside \boxed{}.
\end{verbatim}

\paragraph{Question validity and skill-matching prompt.}
The quality supervisor judges whether a generated question is valid and whether the declared skill matches the question. It does not solve the question, but only returns a binary decision.

\begin{verbatim}
	System:
	You are a strict visual question validity judge. Decide whether the question
	can be answered solely from the provided image and whether the declared skill
	matches the question.
	
	Use the following skill definition and restriction:
	{SKILL_CONTEXT}
	
	Output \boxed{1} only if the question is image-grounded, well-posed,
	answerable from the image alone, and consistent with the declared skill.
	Otherwise, output \boxed{0}.
	
	User:
	[IMAGE]
	Question: {QUESTION}
	Declared Skill: {DECLARED_SKILL}
	
	Judge whether the question is valid and whether the declared skill is correct.
	Do not solve the question. End with \boxed{1} or \boxed{0}.
\end{verbatim}

\paragraph{Answer verification prompt.}
After majority voting over solver responses, the supervisor further verifies whether the candidate pseudo-label is correct.

\begin{verbatim}
	System:
	You are a strict visual question answering verifier. Given an image, a question,
	and a candidate answer, decide whether the candidate answer is correct.
	Output \boxed{1} only if the candidate answer is correct.
	
	User:
	[IMAGE]
	Question: {QUESTION}
	Candidate Answer: {CANDIDATE_ANSWER}
	
	Briefly judge the answer correctness and end with \boxed{1} or \boxed{0}.
\end{verbatim}

\paragraph{Skill context.}
For the skill-matching prompt, \texttt{\{SKILL\_CONTEXT\}} is dynamically replaced by the definition of the declared skill. We use six skill categories: coarse perception, fine-grained perception, instance reasoning, logical reasoning, math \& counting, and science \& technology. For non-math skills, we explicitly forbid questions whose main solution path depends on counting or numerical estimation, so as to reduce skill-label leakage toward the math \& counting category.

\subsection{Overall Training Algorithm}
\label{app:algorithm}
\begin{center}
\begin{minipage}{0.98\linewidth}
\begin{algorithmic}[1]
    \Require Unlabeled image set $\mathcal{X}$, initial VLM $\pi_0$, skill set $\mathcal{K}$, cycle length $b$, number of cycles $C$
    \Ensure Evolved solver $\pi_s$
    \State Initialize questioner and solver with the same VLM: $\pi_q \leftarrow \pi_0$, $\pi_s \leftarrow \pi_0$
    \State Set the quality supervisor as $\pi_{\mathrm{sup}} \equiv \pi_s$
    \State Initialize recent skill statistics $\{n_k\}_{k\in\mathcal{K}}$
    
    \For{$c=1$ to $C$}
        \Statex \textit{// Questioner update}
        \For{$t=1$ to $b$}
            \State Sample images from $\mathcal{X}$ and generate questions $(q,k)$ with $\pi_q$
            \State Estimate question difficulty by sampling multiple responses from $\pi_s$
            \State Use $\pi_{\mathrm{sup}}$ to judge question validity and skill consistency
            \State Compute the questioner reward with difficulty score, validity score, and skill bonus
            \State Update $\pi_q$ with GRPO
        \EndFor
        
        \Statex \textit{// Pseudo-label construction}
        \State Generate candidate questions using the updated questioner $\pi_q$
        \State Sample multiple solver responses and construct majority-voted pseudo-labels
        \State Filter samples by self-consistency, question validity, and answer verification
        \State Apply skill-balanced sampling to obtain the solver training set $\mathcal{D}_s$
        \State Update recent skill statistics $\{n_k\}_{k\in\mathcal{K}}$
        
        \Statex \textit{// Solver update}
        \For{$t=1$ to $b$}
            \State Sample pseudo-labeled data $(x,q,\hat{y},k)$ from $\mathcal{D}_s$
            \State Compute solver rewards by comparing extracted answers with $\hat{y}$
            \State Update $\pi_s$ with GRPO
        \EndFor
        
        \State Refresh the shared supervisor: $\pi_{\mathrm{sup}} \equiv \pi_s$
    \EndFor
    
    \State \Return $\pi_s$
\end{algorithmic}
\end{minipage}
\end{center}
Algorithm above provides the overall training procedure of RISE. We initialize the questioner and solver from the same VLM and use the current solver as a quality supervisor. Each fine-grained alternation cycle consists of three stages: questioner update, pseudo-label construction, and solver update. The questioner is optimized with difficulty, validity, and skill-balancing rewards; the generated VQA samples are then filtered by self-consistency and supervisor verification; and the solver is updated on the resulting skill-balanced pseudo-labeled data. This repeated interaction enables more timely role coordination and more reliable self-evolution.

\section{Qualitative Visualization}
\label{app:qualitative}

In this section, we provide additional qualitative results to further illustrate the failure modes of uncontrolled self-evolution and the behavior of RISE during training. Specifically, we present: (1) representative bad cases without quality and diversity control, (2) the evolution of generated questions for the same image across training stages, (3) examples of questions generated by RISE for different skill types, and (4) the skill distribution statistics of the pseudo-labeled VQA data constructed at different training steps.

\subsection{Bad Cases without Quality and Diversity Control}
\label{app:badcase}

\begin{figure*}[t]
	\centering
	\includegraphics[width=\textwidth]{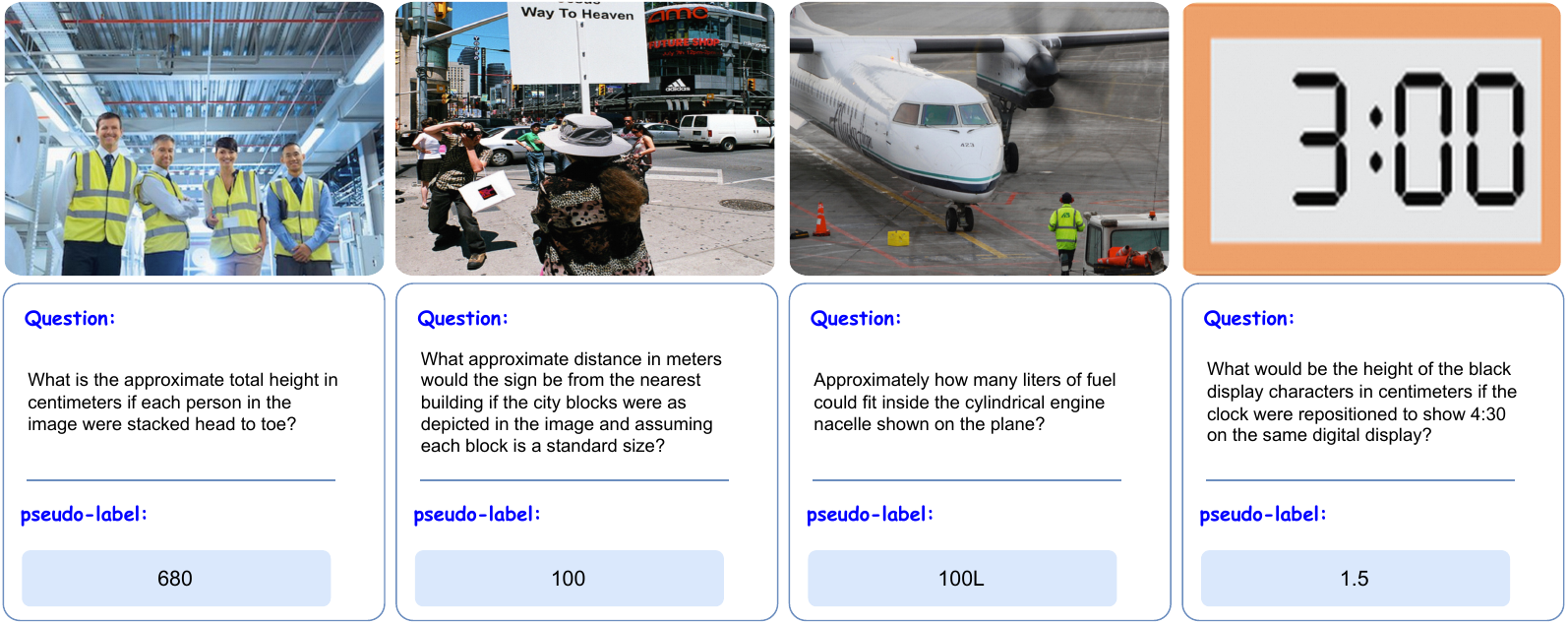}
	\caption{Representative bad cases generated without explicit control of question quality and diversity. Many questions are weakly grounded, ambiguous, or unsuitable as effective supervision. Moreover, the generated content shows a clear collapse toward estimation- and quantity-related questions, reflecting the tendency of difficulty-only optimization to favor uncertain numerical questions.}
	\label{fig:app_badcase}
\end{figure*}

Figure~\ref{fig:app_badcase} shows representative questions generated without explicit control over question quality and diversity. Two observations can be made. First, many generated questions are of poor quality: some are weakly grounded, some are ambiguous or unverifiable, and some are simply unsuitable as effective VQA supervision. In particular, several questions require arbitrary estimation of quantities or physical values from a single image, making them unreliable as training targets. Second, the generated content exhibits a clear collapse in diversity. Instead of covering a broad range of visual skills, the questions become heavily concentrated on numerical estimation or quantity-related reasoning.

This phenomenon is closely related to the difficulty-driven reward used for the questioner. Since the questioner is encouraged to maximize the difficulty score induced by the solver's self-consistency, its objective is effectively to produce questions whose answer consistency is close to 0.5. In practice, estimation-style questions are a convenient shortcut to achieve this target, because they are often uncertain, weakly constrained, and naturally lead to dispersed solver responses. As a result, without additional quality supervision and skill-aware balancing, self-evolution tends to produce low-quality yet reward-seeking questions and gradually collapses to a narrow subset of skills.

\begin{figure*}[h]
	\centering
	\includegraphics[width=\textwidth]{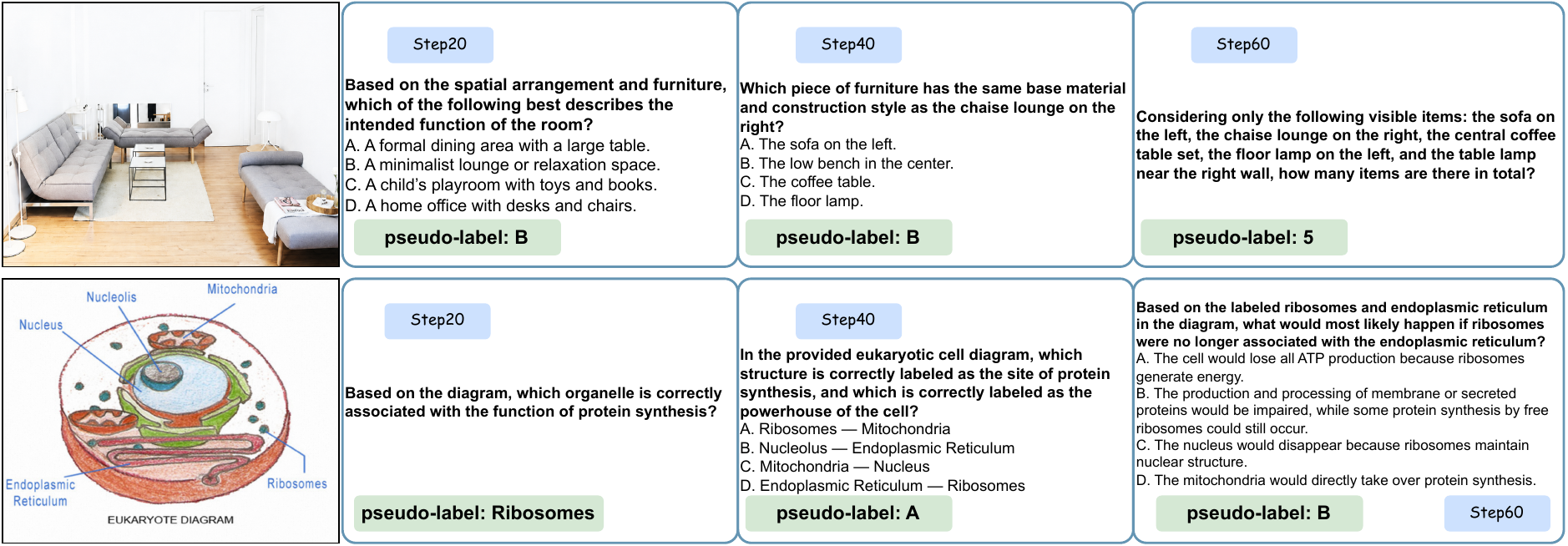}
	\caption{Evolution of generated questions for the same image across different training stages. The generated questions become progressively more challenging over time, while remaining image-grounded and diverse under the guidance of the quality supervisor and skill-aware balancing.}
	\label{fig:app_que_evo}
\end{figure*}
\subsection{Question Evolution on the Same Image}
\label{app:que_evo}

Figure~\ref{fig:app_que_evo} visualizes the questions generated for the same image at different training stages. As training proceeds, the generated questions become progressively more challenging: early-stage questions mainly focus on direct scene understanding or simple recognition, while later-stage questions require more fine-grained analysis, compositional reasoning, or more difficult numerical judgment. This progression indicates that the questioner gradually learns to construct more informative supervision as self-evolution advances.

At the same time, unlike uncontrolled self-evolution, the quality and diversity of generated questions do not collapse. Thanks to the quality supervisor, later questions remain image-grounded, answerable, and semantically valid. Meanwhile, skill-aware dynamic balancing prevents the generation process from collapsing to a single high-reward skill and encourages the questioner to maintain diverse question types during training. Therefore, RISE can increase question difficulty while preserving both question quality and skill diversity.

\subsection{Examples of Generated Questions across Skills}
\label{app:skill_examples}

\begin{figure*}[h]
	\centering
	\includegraphics[width=\textwidth]{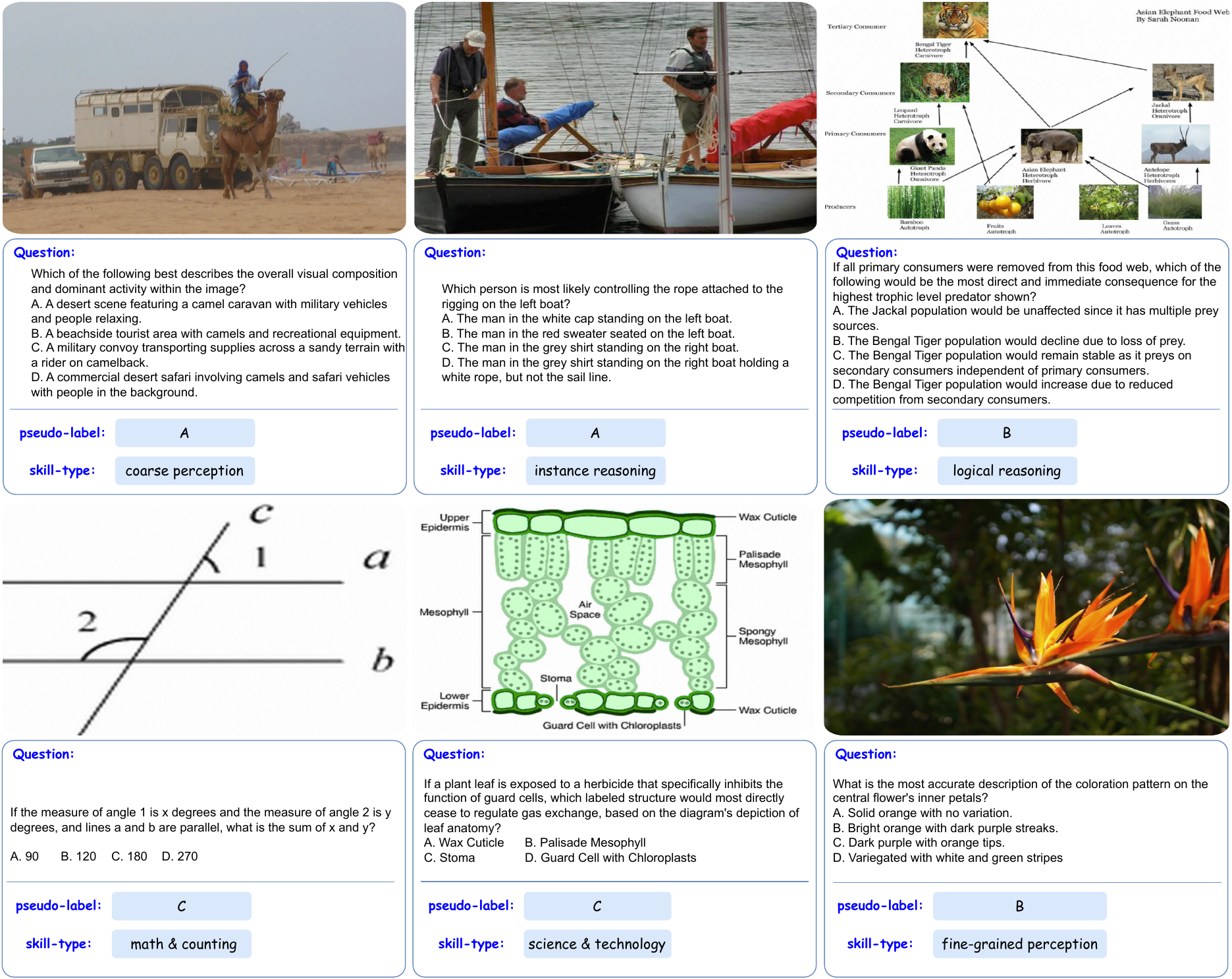}
	\caption{Representative questions generated by RISE for different skill types. The examples show that RISE maintains broad skill coverage, including coarse perception, fine-grained perception, instance reasoning, logical reasoning, math \& counting, and science \& technology.}
	\label{fig:app_skill}
\end{figure*}

Figure~\ref{fig:app_skill} presents representative examples of questions generated by RISE for different skill types. These examples cover multiple categories, including coarse perception, fine-grained perception, instance reasoning, logical reasoning, math \& counting, and science \& technology. The examples qualitatively demonstrate that RISE does not merely generate harder questions, but also maintains broad coverage over different visual and reasoning abilities. This multi-skill coverage is important for preventing mode collapse and for constructing pseudo-labeled data that can support more balanced capability growth of the solver.

\subsection{Skill Distribution of Constructed Pseudo-Labeled Data}
\label{app:skill_sta}

\begin{figure*}[h]
	\centering
	\includegraphics[width=\textwidth]{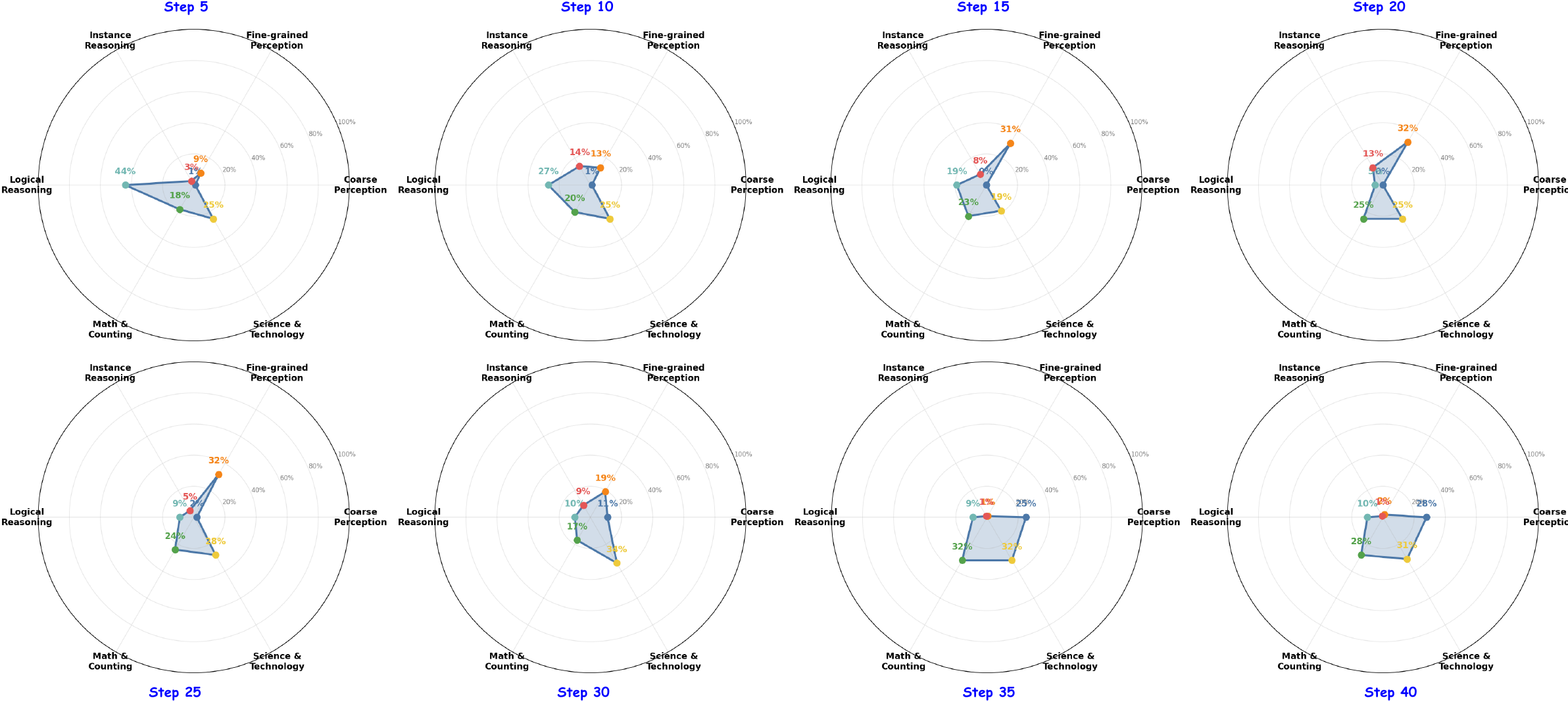}
	\caption{Skill distribution statistics of the pseudo-labeled VQA data constructed by RISE every five training steps. The distributions remain multi-modal rather than collapsing to a single dominant skill, while their relative emphasis changes across stages. Since each round uses 256 randomly sampled images, the observed fluctuations also reflect the interaction between dynamic skill regulation and the sampled image content.}
	\label{fig:app_skill_sta}
\end{figure*}

Figure~\ref{fig:app_skill_sta} shows the skill distribution statistics of the pseudo-labeled VQA data constructed by RISE every five training steps. Overall, the generated data do not collapse to a single dominant skill, which is consistent with our quantitative findings in the main paper. Instead, the skill distribution remains relatively balanced throughout training, while the emphasis on different skills varies across stages.

These variations are expected and meaningful. In each construction round, we generate pseudo-labeled data from a batch of 256 randomly sampled images, so the instantaneous skill distribution is partly influenced by the content of the sampled images. Nevertheless, because the image sampling is random, the persistent multi-skill coverage across different steps indicates that the observed distributions are not caused by fixed dataset bias, but instead reflect the effect of skill-aware dynamic balancing. In other words, RISE dynamically regulates question generation across different skill types, thereby avoiding the severe mode collapse observed in uncontrolled self-evolution.

\begin{figure}[h]
	\centering
	\includegraphics[width=0.8\linewidth]{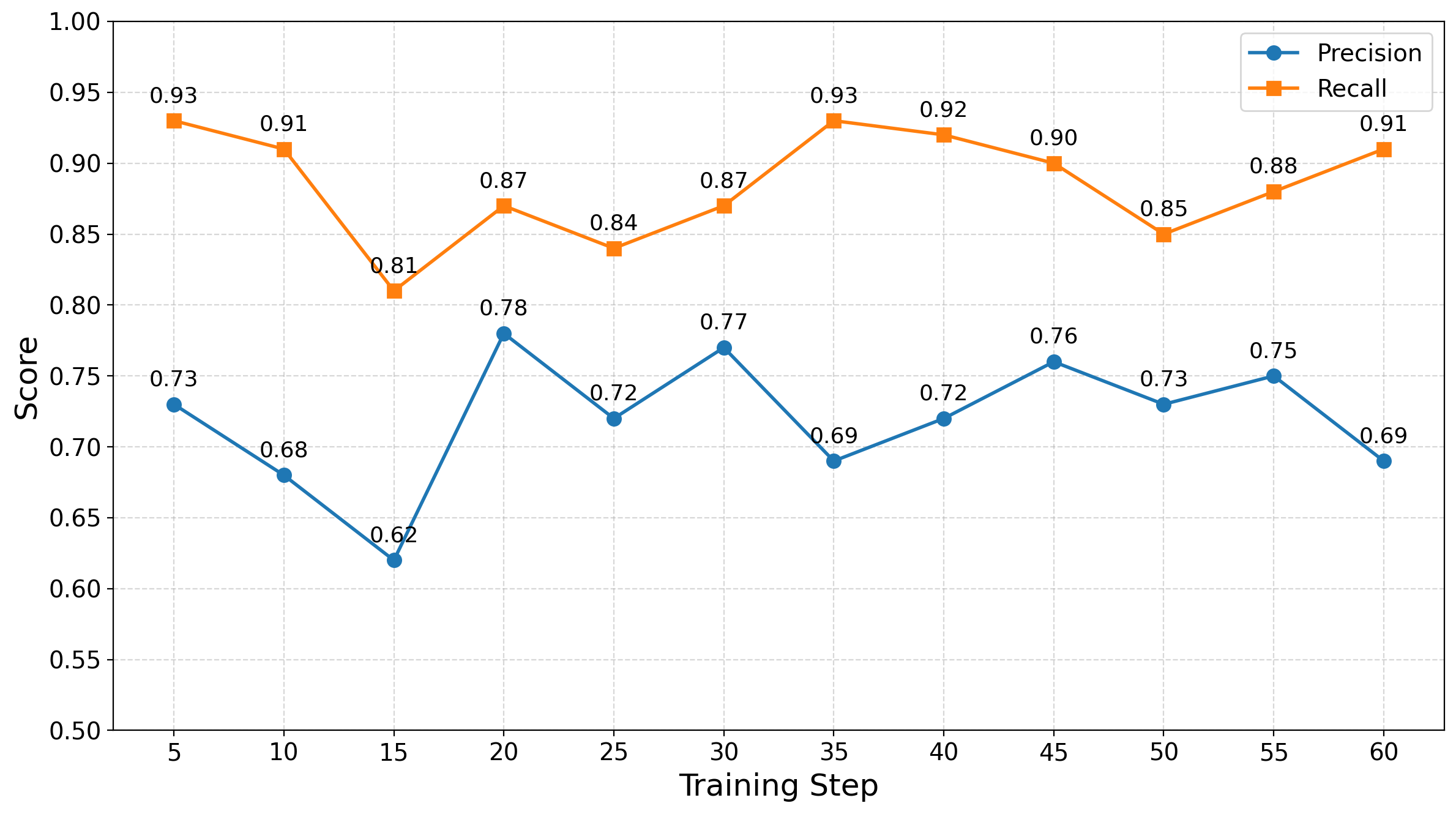}
	\caption{Precision and recall of the supervisor for detecting problematic samples during Qwen3-VL self-evolution. Problematic samples are defined as generated VQA samples that are not valid and correct according to Qwen3.5-Plus. The supervisor maintains consistently high recall across training, while precision is relatively lower but remains stable overall.}
	\label{fig:app_pr_sup}
\end{figure}

\section{Additional Experiments and Analysis}
\label{app:additional_experiments}
\subsection{Analysis of Supervisor Precision and Recall}
\label{app:supervisor_pr}

We further analyze the judgment quality of the supervisor during Qwen3-VL self-evolution. For evaluation, we use Qwen3.5-Plus to assess whether each generated VQA sample is \emph{valid and correct}. Samples that are valid and correct are treated as negative samples, while all remaining samples are treated as problematic positive samples. We then report the precision and recall of the supervisor for detecting these problematic samples.

Figure~\ref{fig:app_pr_sup} shows the precision and recall curves across different self-evolution steps. We can observe that the recall remains consistently high over the whole training process, staying around $0.9$ for most stages. This indicates that the supervisor is effective at identifying and filtering problematic samples during pseudo-label construction, which is important for preventing low-quality questions or unreliable pseudo-labels from entering subsequent solver training.

In contrast, the precision is noticeably lower and fluctuates around $0.7$. This means that, while the supervisor can successfully capture most problematic samples, it also filters out some samples that are actually valid and correct. However, this behavior is acceptable and even desirable in our setting. Since pseudo-labeled data are constructed offline and the generated candidate pool is usually large, our filtering strategy prioritizes high recall over high precision. In other words, it is preferable to remove problematic samples as much as possible, even at the cost of discarding some clean ones, because retaining noisy samples would be more harmful to solver training.

Another important observation is that the supervisor's judgment quality does not show a clear degradation as self-evolution proceeds. In particular, the recall remains stable throughout training, and the precision does not exhibit a monotonic downward trend. This suggests that the supervisor does not suffer from evident reward hacking or progressive collapse during self-evolution, supporting its reliability as a quality-control mechanism in RISE.

\subsection{Why Can Self-Evolution Improve with Imperfect VQA Data?}
\label{app:imperfect_vqa}

Although RISE explicitly improves the quality and diversity of generated VQA data, the constructed pseudo-labeled data are still not perfectly clean. Some generated questions may be ambiguous, weakly grounded, or paired with imperfect pseudo-labels. Nevertheless, as shown in Fig.~\ref{fig:question}, the solver can still obtain consistent performance gains during self-evolution. We provide three possible explanations for this phenomenon.

First, our solver training follows a TTRL-style pseudo-labeling paradigm, where the current policy is improved by using the majority-voted answer from multiple sampled responses, i.e., a Maj@N policy, as the pseudo-label. In this setting, the pseudo-label is not an oracle annotation, but a consensus signal extracted from the model's own response distribution. When the model already has partial knowledge about the task, the majority answer is often more reliable than a single sampled response. Training the current policy toward this consensus can therefore be viewed as a form of self-distillation: it increases the probability of responses that are consistent with the model's dominant correct mode and suppresses unstable minority responses. Although this strategy may occasionally reinforce an incorrect majority answer, it is beneficial on average when the majority signal is more often correct than wrong. In this sense, Maj@N-based training trades off some minority errors for a stronger overall policy, which explains why the model can improve even when the pseudo-labeled data are not perfectly accurate.

Second, even when the final pseudo-label of a generated question is imperfect, the question can still induce useful visual reasoning behavior. In our solver prompt, the model is required to reason step by step before producing the final answer. By inspecting the solver responses to imperfect samples, we find that the model often still performs meaningful intermediate analysis: it identifies relevant visual elements, compares objects, counts instances, reasons about spatial relations, or attempts to connect visual evidence with the question. These intermediate reasoning behaviors can provide useful training pressure, especially when the final answer is not severely misleading. Therefore, self-evolution may improve the solver not only by teaching exact pseudo-labels, but also by repeatedly exercising visual grounding and reasoning procedures. This helps explain why the solver can benefit from generated VQA data even when a subset of answers is noisy or ambiguous.

Third, our experiments show that imperfect data are tolerable, but more reliable data lead to better self-evolution. As shown in Table~\ref{tab:ablation}, removing the supervisor causes the fine-grained self-evolving process to become unstable in later stages, while introducing the quality supervisor substantially stabilizes training. Similarly, Fig.~\ref{fig:question} shows that without explicit quality and diversity control, generated questions can degrade in validity and collapse toward narrow estimation-style categories. These results indicate that noisy pseudo-labels are not harmless: they can limit or even damage self-evolution when they accumulate. The reason RISE achieves stronger and more stable improvement is that the quality supervisor and skill-aware dynamic balancing reduce the proportion of such harmful samples, making the remaining supervision more reliable and diverse.

Overall, the effectiveness of RISE under imperfect pseudo-labels comes from a combination of three factors: Maj@N pseudo-labeling provides a statistically useful consensus signal, step-by-step solver training encourages transferable visual reasoning behavior, and our quality-diversity control mechanisms reduce the amount of harmful supervision. Thus, the presence of some erroneous VQA samples does not contradict the observed performance gains; instead, our results suggest that self-evolution can be robust to moderate pseudo-label noise, while still benefiting substantially from better question quality and broader skill coverage.

\subsection{Ablation on Role Alternation Granularity}
\label{app:alternation_granularity}

We further study the effect of role alternation granularity in RISE. In our fine-grained role alternation, the hyperparameter $b$ denotes the number of update steps in each short alternation cycle. A larger $b$ corresponds to less frequent role switching, while a smaller $b$ makes the interaction between the questioner and the solver more online. In particular, $b=20$ corresponds to the coarse-grained setting used in the main paper, $b=5$ is our default fine-grained setting, and $b=1$ represents an almost fully online-style alternation strategy.

To isolate the effect of alternation granularity, we keep the quality supervisor enabled for all variants but remove the skill-aware dynamic balancing module. This is because skill balancing is itself coupled with the alternation cycle: a smaller $b$ would update the recent skill statistics and apply balancing more frequently, making it difficult to attribute the performance change solely to role alternation granularity. Therefore, all variants in this study use the same quality-control mechanism but do not use skill balancing.

\begin{table}[h]
\centering
\caption{Ablation study on the role alternation granularity $b$ using Qwen2.5-VL-7B-Instruct. All variants use the quality supervisor but remove skill-aware dynamic balancing. $b=20$ denotes the coarse-grained setting, while smaller $b$ values correspond to more frequent role alternation. AVG denotes the sample-size weighted average over all benchmarks.}
\label{tab:granularity_ablation}
\resizebox{\linewidth}{!}{
\begin{tabular}{lcccccccc}
\toprule
Method & MMMU & MMVet & RealWorldQA & ChartQA & MathVerse & MathVision & MathVista & AVG \\
\midrule
Base Model & 49.71 & 51.83 & 57.12 & 79.48 & 40.48 & 23.21 & 58.70 & 48.17 \\
\midrule
$b=1$, Step 20  & 52.28 & 59.17 & 57.91 & 81.20 & 43.73 & 27.47 & 58.80 & 50.94 \\
$b=5$, Step 20  & 53.68 & 57.80 & 58.04 & 81.08 & 42.69 & 27.03 & 59.80 & 50.64 \\
$b=10$, Step 20 & 52.51 & 57.34 & 58.30 & 80.88 & 44.21 & 25.44 & 58.90 & 50.59 \\
$b=20$, Step 20 & 51.23 & 58.26 & 58.56 & 81.56 & 41.83 & 24.51 & 60.40 & 49.81 \\
\midrule
$b=1$, Step 40  & 51.23 & 55.96 & 58.95 & 82.08 & 43.15 & 27.40 & 57.70 & 50.76 \\
$b=5$, Step 40  & 52.86 & 60.09 & 58.82 & 81.64 & 45.15 & 27.21 & 59.90 & 51.65 \\
$b=10$, Step 40 & 54.73 & 52.75 & 59.08 & 81.76 & 44.26 & 27.18 & 57.50 & 51.19 \\
$b=20$, Step 40 & 51.81 & 60.09 & 59.87 & 81.44 & 43.93 & 24.92 & 61.30 & 50.80 \\
\midrule
$b=1$, Step 60  & 52.98 & 54.59 & 60.00 & 81.76 & 42.44 & 26.95 & 59.90 & 50.71 \\
$b=5$, Step 60  & 53.56 & 56.42 & 57.25 & 82.00 & 45.18 & 27.51 & 60.20 & 51.71 \\
$b=10$, Step 60 & 53.79 & 52.29 & 58.30 & 81.32 & 44.21 & 27.47 & 59.30 & 51.17 \\
$b=20$, Step 60 & 54.38 & 57.34 & 58.82 & 81.48 & 44.67 & 23.99 & 60.00 & 50.80 \\
\bottomrule
\end{tabular}
}
\end{table}

As shown in Table~\ref{tab:granularity_ablation}, role alternation granularity has a clear effect on the efficiency of self-evolution. When comparing the settings with $b=5,10,20$, the performance consistently follows the trend of $b=5 > b=10 > b=20$ at the same training step. Specifically, $b=5$ achieves the highest AVG among these three settings at Step 20, Step 40, and Step 60, followed by $b=10$, while the coarse-grained setting $b=20$ performs the worst. This shows that more frequent role alternation consistently improves training efficiency by reducing the feedback lag between the questioner and the solver. This allows the questioner to adapt to a more up-to-date solver and enables the solver to learn from more timely generated data. At Step 20, the fully online-style setting with $b=1$ further achieves the best AVG of 50.94, suggesting that extremely frequent interaction can accelerate early-stage adaptation.

However, the fully online-style setting does not continue to improve in later stages. The AVG of $b=1$ remains almost unchanged from Step 20 to Step 60, changing from 50.94 to 50.76 and 50.71. This suggests that overly frequent alternation may make the interaction between the two roles too reactive. Since the questioner reward depends on the current solver's response consistency, updating the solver after every single step makes the capability boundary highly non-stationary. As a result, the questioner may overfit to transient weaknesses of the solver rather than constructing a stable and progressively informative curriculum. Meanwhile, the solver also has less opportunity to consolidate the pseudo-labeled data generated under the current questioner before the data distribution changes again.

In contrast, the moderate fine-grained setting $b=5$ achieves the best long-term performance, reaching 51.65 at Step 40 and 51.71 at Step 60. This shows that $b=5$ provides a better trade-off between feedback freshness and training stability. Compared with the coarse-grained setting $b=20$, it substantially shortens the feedback loop; compared with the fully online setting $b=1$, it still preserves a short but stable phase for the questioner to adapt and for the solver to consolidate. These results support our choice of $b=5$ as the default fine-grained alternation granularity in RISE.

\subsection{Additional Ablation Study on Qwen3-VL}
\label{app:qwen3_ablation}

We further conduct an ablation study on Qwen3-VL-8B-Instruct to examine whether the contribution of each component remains consistent on a stronger backbone. The dataset order follows the main paper, and we report the results at Step 60. Here, ``w/o Skill'' denotes removing the skill-aware dynamic balancing module, ``w/o Sup.'' denotes removing the quality supervisor, and ``w/o Fine-grained'' denotes replacing fine-grained role alternation with coarse-grained alternation.

\begin{table}[h]
	\centering
	\caption{Additional ablation study on Qwen3-VL-8B-Instruct at Step 60. AVG denotes the sample-size weighted average over all benchmarks.}
	\label{tab:qwen3_ablation}
	\resizebox{\linewidth}{!}{
		\begin{tabular}{lcccccccc}
			\toprule
			\textbf{Method} & \textbf{MMMU} & \textbf{MMVet} & \textbf{RealWQA} & \textbf{ChartQA} & \textbf{MathVerse} & \textbf{MathVision} & \textbf{MathVista} & \textbf{AVG} \\
			\midrule
			RISE & 59.63 & 62.84 & 70.33 & 83.64 & 49.11 & 36.10 & 68.70 & 57.37 \\
			RISE w/o Skill & 56.94 & 63.30 & 69.02 & 83.32 & 49.52 & 32.95 & 67.40 & 56.36 \\
			RISE w/o Sup. & 56.71 & 62.84 & 68.89 & 82.92 & 49.47 & 32.65 & 67.80 & 56.19 \\
			RISE w/o Fine-grained & 56.13 & 60.55 & 68.50 & 82.72 & 48.12 & 32.95 & 67.60 & 55.65 \\
			\bottomrule
		\end{tabular}
	}
\end{table}

As shown in Table~\ref{tab:qwen3_ablation}, removing any component leads to a lower overall weighted average, confirming that the three designs all contribute to the final performance on Qwen3-VL-8B-Instruct. Compared with the full RISE model, removing skill-aware dynamic balancing, the quality supervisor, and fine-grained role alternation decreases AVG by 1.01, 1.18, and 1.72 points, respectively.

It is worth noting that, unlike the Qwen2.5-VL ablation in the main paper, removing the supervisor on Qwen3-VL does not cause a severe late-stage collapse. We attribute this difference to the stronger base capability of Qwen3-VL. As shown in Fig.~\ref{fig:question}, although the quality of generated questions still decreases without the supervisor, Qwen3-VL can retain a non-trivial proportion of valid-and-correct samples in later stages, with the final ratio remaining around 30\%. In contrast, Qwen2.5-VL drops to below 10\% in the late stage, making the accumulated noisy supervision much more harmful. Therefore, the effect of the supervisor is closely related to the base model's ability to generate and filter reliable self-evolving data: stronger models are more robust to quality degradation, but the supervisor still provides clear gains by improving the reliability of pseudo-labeled supervision.

\section{Limitations}
\label{sec:limitations}

Although RISE substantially improves the reliability of VLM self-evolution from unlabeled images, several limitations remain. First, we observe that self-evolution still shows a convergence trend in later stages. While RISE brings consistent gains across training steps, the marginal improvement gradually becomes smaller, suggesting that majority-voted pseudo-label supervision may provide diminishing learning signals as the solver becomes stronger. Second, the behavior of self-evolution varies across different base models. Models with different initial capabilities may exhibit different question-generation distributions, different levels of VQA data quality, and different degrees of robustness to noisy pseudo-labels. A deeper understanding of how base model capability affects the dynamics of self-evolution remains an important direction for future work. Third, the current fine-grained role alternation frequency is controlled by manually selected hyperparameters. Although this simple strategy is effective, a more adaptive mechanism that automatically adjusts the alternation frequency according to the current capability and stability of the model could further improve the efficiency and reliability of self-evolution.

\section{Broader Impacts}

RISE seeks to reduce the dependence of VLM post-training on costly human-annotated multimodal supervision by enabling models to improve from unlabeled images and self-generated learning signals. This may improve the scalability and accessibility of multimodal model development, especially for visual reasoning tasks where high-quality annotations are expensive to obtain.

However, self-evolving VLMs may also amplify biases, hallucinations, or visual grounding errors inherited from the base model, since the training signals are partially generated and verified by the model itself. Improved VLM capabilities may also be misused in high-stakes or privacy-sensitive applications, such as automated surveillance or misleading visual interpretation. Although RISE introduces quality supervision and skill-aware balancing to mitigate unreliable supervision, these mechanisms do not remove all risks. Careful evaluation is still necessary before deploying evolved models in real-world scenarios.

\newpage

\end{document}